\documentclass[letterpaper, 10 pt, conference]{ieeeconf}  

\IEEEoverridecommandlockouts                              

\usepackage{xcolor}

\usepackage{graphicx}

\usepackage{gensymb}
\usepackage[hyphens]{url}
\usepackage{multirow}
\usepackage{multicol}
\usepackage{tabularx}
\usepackage{dsfont}
\usepackage{url}
\usepackage{color}
\usepackage{subcaption}
\usepackage{caption}
\usepackage{mathrsfs}
\usepackage{float}
\usepackage{booktabs}
\usepackage{dcolumn}
\usepackage{dingbat}
\usepackage{setspace}
\let\checkmark\undefined
\usepackage{amssymb}
\usepackage{amsmath}
\usepackage[linesnumbered,ruled,vlined]{algorithm2e}
\usepackage[colorlinks,linkcolor=red,anchorcolor=blue,citecolor=green]{hyperref}

\usepackage{xcolor}

\title{\LARGE \bf
Robust Collaborative Perception without External \\
Localization and Clock Devices
}

\author{Zixing Lei$^{1}$, Zhenyang Ni$^{1}$, Ruize Han$^{2}$, Shuo Tang$^{1}$, Dingju Wang$^{1}$, Chen Feng$^{3}$, Siheng Chen$^{1, 4}$, Yanfeng Wang$^{4,1}$
\thanks{$^{1}$Cooperative Medianet Innovation Center , Shanghai Jiao Tong University, China. 
        {\tt\small \{chezacarss, 0107nzy, tanshu, sihengc, wangyanfeng622\}@sjtu.edu.cn.}}%
\thanks{$^{2}$Shenzhen Institute of Advanced Technology, Chinese Academy of Sciences
        {\tt\small rz.han@siat.ac.cn}}%
\thanks{$^{3}$New York University, USA. {\tt\small cfeng@nyu.edu}}%
\thanks{$^{4}$Shanghai AI laboratory, China.}%
}

\begin{document}

\maketitle
\thispagestyle{empty}
\pagestyle{empty}

\begin{abstract}
A consistent spatial-temporal coordination across multiple agents is fundamental for collaborative perception, which seeks to improve perception abilities through information exchange among agents. 
To achieve this spatial-temporal alignment, traditional methods depend on external devices to provide localization and clock signals. 
However, hardware-generated signals could be vulnerable to noise and potentially malicious attack, jeopardizing the precision of spatial-temporal alignment. 
Rather than relying on external hardwares, this work proposes a novel approach: aligning by recognizing the inherent geometric patterns within the perceptual data of various agents. 
Following this spirit, we propose a robust collaborative perception system that operates independently of external localization and clock devices. 
The key module of our system,~\emph{FreeAlign}, constructs a salient object graph for each agent based on its detected boxes and uses a graph neural network to identify common subgraphs between agents, leading to accurate relative pose and time. 
We validate \emph{FreeAlign} on both real-world and simulated datasets. The results show that, the ~\emph{FreeAlign} empowered robust collaborative perception system perform comparably to systems relying on precise localization and clock devices. \href{https://github.com/MediaBrain-SJTU/FreeAlign}{Code} will be released.

\end{abstract}


\section{INTRODUCTION}
Collaborative perception~\cite{v2vnet, v2xsim} facilitates the sharing of complementary perceptual information among multiple agents~\cite{opv2v}, fostering a comprehensive understanding of their surroundings~\cite{dairv2x}. This approach offers a means to address several inherent limitations of single-agent perception, including occlusion and long-range challenges.  Previous methods~\cite{opencda, v2xvit, v2xseq} provides efficacious collaboration techniques and promising performance~\cite{cobevt,poseerror}. Collaborative perception has the potential to be widely used in vehicle-to-everything communication-assisted autonomous driving\cite{disconet,where2comm, 3dpc} and multi-UAVs (unmanned aerial vehicles)~\cite{aerial, multiuav, marluav}.
\begin{figure}[t]
    \centering{\includegraphics[width=0.47\textwidth]{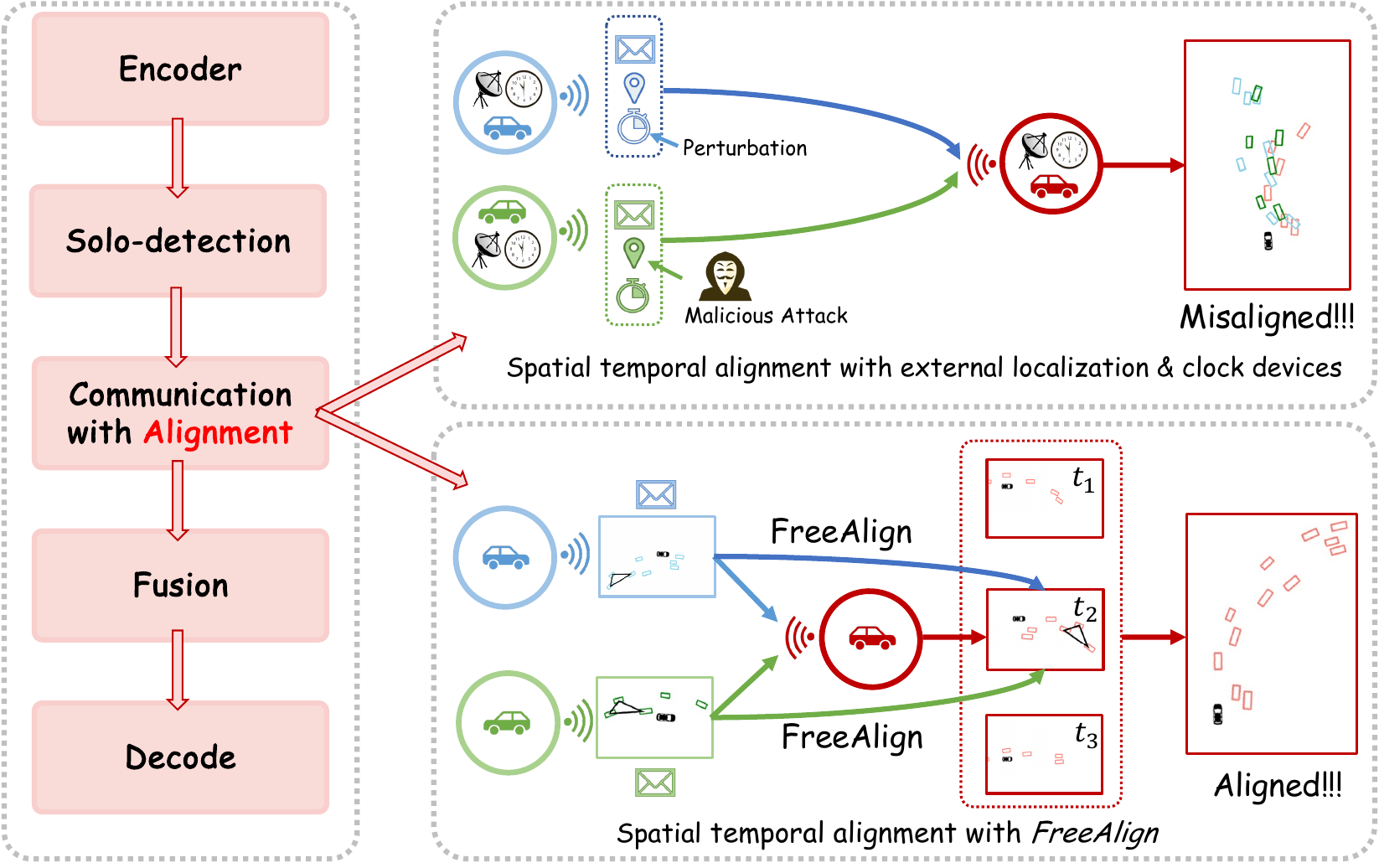}}
    \caption{An illustration of collaborative perception in autonomous driving. The upper side shows in traditional approach, inaccurate localization and clock signal mislead the collaborative perception system. The lower side show the proposed \emph{FreeAlign} enables robust collaborative perception that operates without external devices for localization or synchronized clock. The core of FreeAlign is to associate the same objects perceived by multiple agents based on similar geometric structures among those objects.}
    \label{illu}
    \vspace{-5mm}
\end{figure}
\begin{table*}[t]
\centering
\scriptsize
\caption{The notations and relation of the important spatial-temporal concepts in collaborative perception.}
\begin{tabular}{c|l}
\hline
Notations & \multicolumn{1}{c}{Paraphrase} \\ \hline
$t_i$ & Current time of ego agent $i$. Following notations are based on the clock in the agent $i$.\\
$t_{j \rightarrow i}/ \widetilde{t}_{j \rightarrow i}$ & Accurate/Estimated(Gotten by external devices) time of collaborative message sent from agent $j$ to ego agent $i$ was perceived.\\
$\Delta t_{j \rightarrow i}/ \widetilde{\Delta t}_{j \rightarrow i}$& Accurate/Estimated latency of the collaborative message sent from agent $j$ to agent $i$, $\Delta t_{j \rightarrow i} = t_i - t_{j \rightarrow i}, \widetilde{\Delta t}_{j \rightarrow i} = t_i - \widetilde{t}_{j \rightarrow i}$.\\
$\delta t_{j \rightarrow i}$ & Clock deviation between agent $i$ and agent $j$. $\delta t_{j \rightarrow i} =t_{j \rightarrow i} - \widetilde{t}_{j \rightarrow i} = \Delta t_{j \rightarrow i} - \widetilde{\Delta t}_{j \rightarrow i}$. $\delta t_j$ will harm the compensation module.\\
$\tau$ & Sample interval in collaborative perception system($\leq 100$ms)\\
$\xi_i^{t_i}/\widetilde{\xi}^{t_i}_i$ & Accurate/Estimated(Gotten by external devices) pose of agent $i$ at time $t_i$ in the global coordinate system. \\
$\Delta{\xi}_{j \rightarrow i}^{t_{j \rightarrow i}}/\widetilde{\Delta{\xi}}_{j \rightarrow i}^{t_{j \rightarrow i}}$ & Accurate/Estimated relative pose between agent $j$ at time $t_{j \rightarrow i}$ and agent $i$ at time $t_i$, $\Delta{\xi}_{j \rightarrow i}^{t_{j \rightarrow i}} = f_{\text{trans}}\left(\xi_i^{t_i}, \xi_j^{t_{j \rightarrow i}}\right)$. \\  
$\delta \xi^{t_{j\rightarrow i}}_{j \rightarrow i}$ & Relative pose error between agent $j$ at time $t_{j \rightarrow i}$ and agent $i$ at time $t_i$, $\delta \xi^{t_{j\rightarrow i}}_{j \rightarrow i} = \Delta{\xi}_{j \rightarrow i}^{t_{j \rightarrow i}} - \widetilde{\Delta{\xi}}_{j \rightarrow i}^{t_{j \rightarrow i}} =(\delta x^{t_{j\rightarrow i}}_{j \rightarrow i},\delta y^{t_{j\rightarrow i}}_{j \rightarrow i},\delta \theta^{t_{j\rightarrow i}}_{j \rightarrow i}) $.\\
$\mathcal{O}_{i}$ & Odometry of agent $i$. Used to calculate the relative pose among agent $i$'s observations at different times.\\
\hline
\end{tabular}
\vspace{-6mm}
\label{notations}
\end{table*}
A consistent spatial-temporal coordinate system accessible to all agents is the cornerstone of collaborative perception. Traditional methods leverage high-end external devices, like GPS+RTK receivers and synchronized global clocks, to log poses and timestamps, ensuring accurate spatial-temporal alignment. However, this hardware-based manner introduces three significant issues.
\textbf{Global Localization Noise}: The presence of noise in device outputs(such as GPS) causes
inaccurate relative transformation between collaborative messages, 
leading to performance degradation. While current state-of-the-art practices~\cite{coalign} tried to address this issue, it can only manage minor noise, such as translations less than $1m$, and struggles with more serious noise.
\textbf{Clock Deviation}: Previous methods such as \cite{dairv2x, syncnet} for collaborative perception often work under the presumption of known latency. Yet, the latency is calculated with timestamps based on clocks of different agents. Therefore, deviation across agents' clocks may produce inaccurate latency measurements, leading to temporal alignment errors.
\textbf{Vulnerability to Malicious Attacks}: The reliance on external spatial-temporal data, presenting an additional avenue for illusory data injections, makes the system more susceptible to disruptions from malicious attacks, especially through V2V network\cite{attack}.

In this work, we propose a robust collaborative perception system that operates independently of external devices for localization and clock. Its key module, \emph{FreeAlign}, is a novel spatial-temporal alignment method that leverages graph matching techniques to identify similar geometric patterns within the perceptual data of various agents, ensuring accurate alignment in both spatial and temporal domains; see an illustration in Fig. \ref{illu}.~\emph{FreeAlign} comprises three key components: i) salient-object graph learning, which uses a Graph Neural Network (GNN) to capture comprehensive edge features among the salient objects detected by each agent; ii) multi-anchor-based subgraph searching, which identifies the proximate maximum common subgraph across two salient-object graphs;  iii) relative transformation calculation, which leverages the common subgraph to calculate the relative pose and latency between two collaborative messages. 

The advantages of our system are twofold: i) it provides a machine learning approach to substitute global localization and synchronized devices, substantially bolstering the robustness of collaborative perception; and ii) the key component \emph{FreeAlign}, allows for seamless integration with numerous established methods to compose our system without necessitating retraining of the collaborative perception architecture.

In summary, the main contribution of this work are:
\begin{list}{$\bullet$}{\setlength{\itemindent}{1mm}\setlength{\leftmargin}{1mm}}
\item We propose the first collaborative perception system without relying on external localization and clock devices;
\item We propose \emph{FreeAlign}, a novel graph-matching-based spatial-temporal alignment method that can be seamlessly integrated into existing collaborative perception systems.
\item We conduct extensive experiments for collaborative LiDAR-based object detection in simulated and real-world datasets to validate that, integrating \emph{FreeAlign} leads to considerable improvements for a majority of previous methods when facing the typical issues.
\end{list}

\section{Related Work}
\textbf{Collaborative Perception.}
The rise of multi-agent collaborative perception addresses limitations inherent to single-agent perception, such as occlusion. Several datasets, including V2X-Sim\cite{v2xsim}, OPV2V\cite{opv2v}, and DAIR-V2X\cite{dairv2x}, contribute to the body of research. Leveraging these datasets, a host of methods have emerged. For instance, V2VNet\cite{v2vnet} integrates collaboration into perception and prediction. Studies such as Who2comm~\cite{who2com}, When2comm~\cite{when2com}, and Where2comm\cite{where2comm} optimize perception performance against bandwidth efficiency, while SyncNet~\cite{syncnet} addresses communication latency. However, these approaches often assume reliable input from localization and clock devices, assuming negligible noise interference. In contrast, our work envisions collaborative perception independent of such external devices.

\textbf{Graph Matching.}
Graph matching and subgraph search processes are pivotal in various domains, and this study uses them for spatial-temporal alignment. Algorithms like VF2\cite{vf2}, QuickSI\cite{quicksi}, VF2x\cite{vf2x}, VF2 Plus\cite{vf2plus}, and VF3\cite{vf3} employ feasibility functions and depth-first search for graph isomorphism challenges, although they have limitations with graphs of varying node numbers. SuperGlue\cite{superglue} utilizes a neural framework to match local features through a differentiable optimal transport solution. PCA-GM\cite{pcagm} focuses on node embeddings and affinity metrics, but can overfit under certain conditions. These existing methods often falter in collaborative perception due to an over-reliance on visual features and pixel coordination. In contrast, our proposed approach tailors specifically to collaborative perception, emphasizing invariant geometric interrelations. For effective feature extraction, we adopt graph representation learning, incorporating techniques like node2vec\cite{node2vec}, deepwalk\cite{deepwalk}, and GNN methods such as GCN\cite{GCN}, GAT\cite{GAT}, GATv2\cite{GATV2}, and EGAT\cite{EGAT}. Particularly, our implementation utilizes EGAT, which accentuates edge features for enhanced subgraph search reliability.

\begin{figure*}
    \centering
    \includegraphics[width=0.95\textwidth]{ 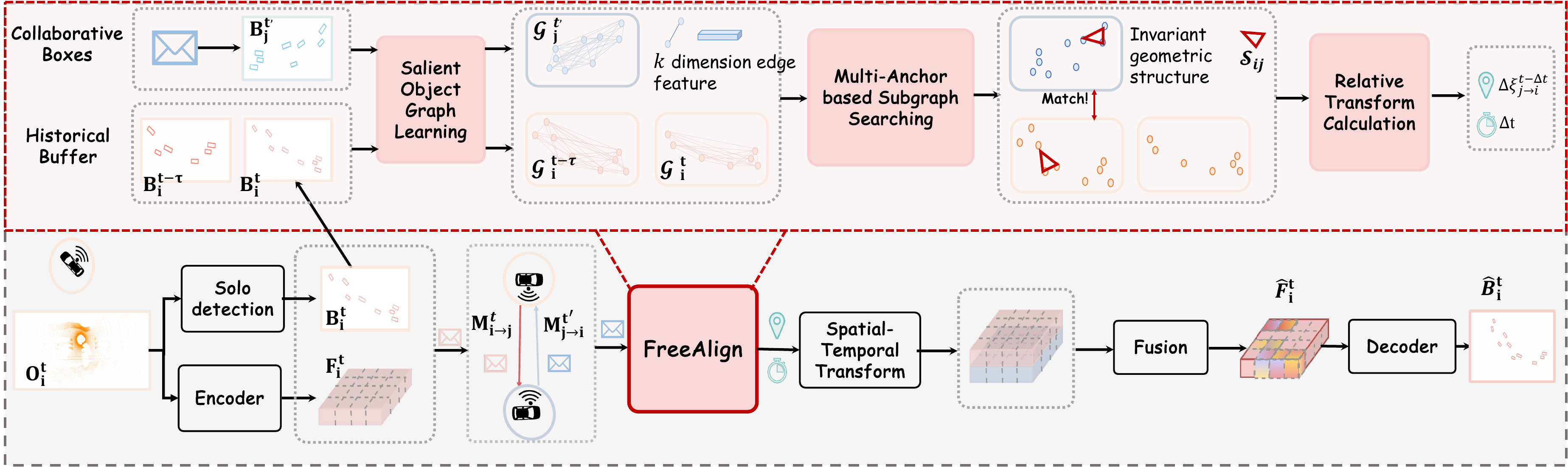}
    \caption{Overview of the proposed robust collaborative perception framework. The key module is \emph{FreeAlign}, which leverages salient-object graphs to achieve spatial-temporal alignment.
    }
    \vspace{-5mm}
    \label{system}
\end{figure*}

\section{Collaborative Perception System}
Consider $N$ agents collaboratively detecting 3D objects within the scene.  Let $\mathbf{X}^t_i$ be the observation of the $i$-th agent at time $t$, $\mathbf{B}_i^t$ and $\widehat{\mathbf{B}}^t_i$ be the detected boxes before and after collaboration, respectively. For the $i$-th agent, the proposed collaborative 3D object detection works as follows:
\begin{subequations}
\label{system}
\small
\begin{align}
& \hspace{-3mm}
\mathbf{F}_i^{t_i}, \mathbf{B}^{t_i}_i=f_{\text {solo-detection}}\left(\mathbf{X}^{t_i}_i\right)
\label{encoder}, \\
& \hspace{-4mm}
\widetilde{\Delta{\xi}}_{j \rightarrow i}^{t_{j \rightarrow i}}, \widetilde{\Delta t}_{j \rightarrow i} = f_{\text{FreeAlign}}\left(\left\{\mathbf{B}^{t_i-k\tau}_i\right\}_{k=0,1,\cdots},\mathbf{B}^{t_{j \rightarrow i}}_j, \mathcal{O}_{i}\right),
\label{1b}\\
& \hspace{-3mm}\widetilde{\mathbf{F}}^{t_{j\rightarrow i}}_{j \rightarrow i}, \widetilde{\mathbf{B}}^{t_{j\rightarrow i}}_{j \rightarrow i} = f_{\text {spatial-trans}}\left(\widetilde{\Delta{\xi}}_{j \rightarrow i}^{t_{j \rightarrow i}}, \mathbf{F}^{t_{j \rightarrow i}}_j, \mathbf{B}^{t_{j \rightarrow i}}_j\right), 
\label{transform}\\
& \hspace{-3mm}
\widetilde{\mathbf{F}}^{t_i}_{j \rightarrow i} =f_{\text {tempoal-trans}}\left(\left\{\widetilde{\Delta t}_{j \rightarrow i}, {\widetilde{\mathbf{F}}}^{t_{j\rightarrow i}-k\tau}_{j \rightarrow i}\right\}_{k = 0, 1,\cdots}\right), 
\label{compensation}\\
& \hspace{-3mm}
\widehat{\mathbf{F}}_i^{t_i}=f_{\text {fusion}}\left(\mathbf{F}^{t_i}_i,\left\{\widetilde{\mathbf{F}}^{t_i}_{j \rightarrow i}\right\} \right)_{j\in \mathcal{N}_i}, 
\label{fusion}\\
& \hspace{-3mm}
\widehat{\mathbf{B}}^{t_i}_i=f_{\text {decoder}}\left(\widehat{\mathbf{F}}_i^{t_i}\right),\label{decoder}
\end{align}
\label{wholesystem}
\end{subequations}
where $\mathbf{F}_i^{t}$ and $\widehat{\mathbf{F}}_i^{t}$ are  the $i$-th agent's feature maps before and after collaboration at time $t$, respectively, $\widetilde{\mathbf{F}}^{t}_{j \rightarrow i}$ is the feature map transmitted from Agent $j$ to $i$ at time $t$, $\widetilde{\Delta{\xi}}_{j \rightarrow i}^{t_{j \rightarrow i}}$ and $\widetilde{\Delta t}_{j \rightarrow i}$ are the relative pose and time between Agents $i$ and $j$, and $\mathcal{N}_i$ is the collaborator of the $i$-th agent.
 
In Step~\eqref{encoder}, Agent \(i\) encodes its observed data \(\mathbf{X}^{t_i}_i\) into a BEV feature map \(\mathbf{F}_i^{t_i}\), thereafter producing detected boxes \(\mathbf{B}^{t_i}_i\). The agent subsequently exchanges this information with its collaborators \(\mathcal{N}_i\). Notably, the exact time \(t_{j\rightarrow i}\) remains \textbf{unknown} to agent \(i\), while the estimated time \(\widetilde{t}_{j\rightarrow i}\) is noisy due to clock deviations.
In Step~\eqref{1b}, Agent \(i\) computes the relative pose and latency between its historical observations and each of its collaborators by utilizing proposed \emph{FreeAlign}, achieving spatial-temporal alignment. 
In Step~\eqref{transform}, Agent \(i\) transforms the $\mathbf{F}^{t_{j \rightarrow i}}_j, \mathbf{B}^{t_{j \rightarrow i}}_j$ to its own spatial coordinates at time \(t_i\), resulting in $\widetilde{\mathbf{F}}^{t_{j\rightarrow i}}_{j \rightarrow i}, \widetilde{\mathbf{B}}^{t_{j\rightarrow i}}_{j \rightarrow i}$. Note that $\widetilde{\mathbf{F}}^{t_{j\rightarrow i}}_{j \rightarrow i}, \widetilde{\mathbf{B}}^{t_{j\rightarrow i}}_{j \rightarrow i}$ are noisy since $\widetilde{\Delta{\xi}}_{j \rightarrow i}$ is estimated. 
In Step~\eqref{compensation}, Agent \(i\) compensates for potential noisy latencies using past collaborative feature maps. 
In Step~\eqref{fusion}, Agent \(i\) aggregates all compensated feature maps to yield a consolidated feature map \(\widehat{\mathbf{F}}_i^{t_i}\), which generates the final detections \(\widehat{\mathbf{B}}^{t_i}_i\) in Step~\eqref{decoder}. 

Precise spatial-temporal alignment is the foundation of effective collaborative perception. A misalignment in this step would significantly damage the subsequent procedures, including feature transformation, fusion and collaborative detection. Traditionally, to achieve the spatial-temporal alignment as Step.~\eqref{1b}, each agent needs an external localization and clock devices to receive signals and provide its globally synchronized pose and time, subsequently computing relative transformations between collaborative messages. Due to the dependence of noisy external signals, this hardware-based approach could cause numerous issues, such as localization noise, clock deviation and vulnerability to malicious attacks. Our proposed method, \emph{FreeAlign}, specifically addresses this by emphasizing perceptual data, utilizing graph matching and machine learning for alignment. A detailed discussion on this will be presented in the following section.


\section{FreeAlign: Graph matching based Spatial-Temporal Alignment} 

To reform the traditional approach, our intuition is that the geometric relation of shared objects (in the real world) between two collaborative messages is unified if both messages are perceived simultaneously, thereby allowing spatial alignment according to the geometric relation and latency determination through a time-series search.
Motivated by this, we present \emph{FreeAlign}, a graph matching based spatial-temporal alignment approach. \emph{FreeAlign} adeptly discerns the consistent geometric structure across collaborative messages, facilitating accurate computation of relative transformation in the spatial-temporal continuum. 
This method comprises three essential modules, i.e., graph learning, subgraph searching, and transformation calculation, which are presented as below.



\subsection{Salient-Object Graph Learning}
\label{subsection:Salient-Object Graph Learning}

To identify similar geometric structures between two collaboration messages, a metric for assessing similarity is required. An intuitive approach is to use the attributes of the objects detected by agents. 
Being invariant to coordinate systems and more distinctive, relative distance serves as a superior metric. However, sole reliance on it can result in confusion due to noise and coincidental matches. To refine the attribute for matching, we propose to model all boxes perceived by one agent as a salient-object graph, where each node is one box. A GNN can then be employed to learn edge features through message passing, enhancing their comprehensiveness and expressiveness by incorporating information from both relative distance and geometric structure. This, in turn, facilitates a more effective identification process.


For the $i$-th agent at timestamp $t$,  let $\mathcal{G}_{i}^t=\left(\mathcal{V}_{i}^t, \mathcal{E}_{i}^t \right)$ be a fully-connected salient-object graph, where $\mathcal{V}_{i}^t$ is the set of $n$ nodes with each node one salient object detected by agent $i$ and $\mathcal{E}_{i}^t$ is the edge set modeling all the pairwise relationships between salient objects. Here we aim to learn a tensor of edge features $\mathbf{W}_{i}^t \in \mathbb{R}^{n \times n \times k}$, where each pillar is the edge feature $(\mathbf{W}_{i}^t)_{pq} \in \mathbb{R}^{k}$ that reflects the invariant proximity between two salient objects $p,q$. Here the invariance means that whatever an agent's pose is, the edge feature of two salient objects is unaltered. This implies that when both agents $i$ and $j$ detect the same two salient objects at the same time $t$, the corresponding edge features in $\mathbf{W}_{i}^t$ and $\mathbf{W}_{j}^t$ would be the same. 
Let $\mathbf{R}_i^t \in \mathbb{R}^{n \times n}$ is a matrix that encodes the relative distance between each pair of nodes, to learn such an edge feature, we use EdgeGAT~\cite{EGAT} as the graph representation learning model, that is,
$\mathbf{W}_i^t = f_{\text{EdgeGAT}}\left(\mathcal{G}_i^t, \mathbf{R}_i^t\right).$ 

For the training of the EdgeGAT model, we employ a contrastive learning loss which can be formulated as follows
\begin{equation}
\small
    \label{loss}
        \begin{aligned}
            L = &\frac{1}{\left|\mathcal{M}\right|}\sum_{\left(p,q\right) \sim \mathcal{M}} \left\Vert \left(\mathbf{W}_{i}^t\right)_{pq}-\left(\mathbf{W}_j^{t'}\right)_{ef} \right\Vert_2 + \\
            	&\frac{1}{\left|\mathcal{U}\right|}\sum_{\left(u,v\right) \sim \mathcal{U}} \text{max} \left(\gamma-\left\Vert \left(\mathbf{W}_{i}^t\right)_{pq}-\left(\mathbf{W}_j^{t^\prime}\right)_{uv}\right\Vert_2, 0 \right),
        \end{aligned}
\end{equation}
where $\left(\mathbf{W}_{i}^t\right)_{pq}$  and $\left(\mathbf{W}_j^{t^\prime}\right)_{ef}$ represent the edge feature  between nodes $p$ and $q$ in $\mathcal{G}^t_i$, and $e$ and $f$ in $\mathcal{G}^{t^\prime}_j$, respectively. 
$(p,q) $ denotes an edge in the matched edge set $\mathcal{M}$ and edge $(p,q)$ matches edge $(e,f)$.
$(u,v) $ denotes an edge in the unmatched edge set $\mathcal{U}$, which is sampled from all edges without match.
$\gamma$ serves as the margin.  This contrastive loss aids the GNN in learning distinctive edge features that facilitate the identification of matches between two graphs.
\begin{figure*}
    \centering
    \includegraphics[width=0.93\linewidth]{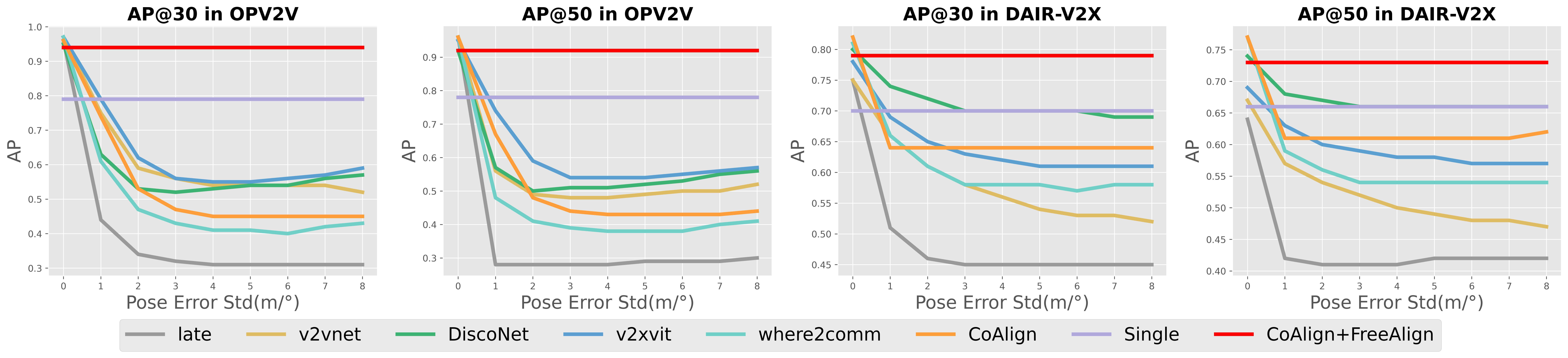}
  \captionof{figure}{Detection performance on OPV2V \cite{opv2v} and DAIR-V2X\cite{dairv2x} datasets with pose noises following Gaussian distribution in the testing phases. The performance of baselines drops significantly as noise increases, while \emph{FreeAlign}'s result is stable.}
  \label{fig:sptial_result}
\end{figure*}

\begin{table*}[]
   \centering
\tiny
\caption{Comparision between \emph{FreeAlign} and other matching methods. Only \emph{FreeAlign} can achieve accurate matching in the collaborative perception scenarios.}\vspace{-1mm}
\resizebox*{0.9\linewidth}{!}{
\begin{tabular}{l|ccc|ccc}
\hline
          & \multicolumn{3}{c|}{OPV2V}                        & \multicolumn{3}{c}{DAIR-V2X}                         \\ \cline{2-7}
          & $\delta \theta$($^\circ$)       & ${\sqrt{({\delta x})^{2}+({\delta y})^{2}}}$(m) & Error rate     & $\delta \theta$ ($^\circ$)        & ${\sqrt{({\delta x})^{2}+({\delta y})^{2}}}$(m) & Error rate     \\ \hline
ICP\cite{icp}       & 61.91          & 24.75          & 93.90\%         & 79.82            & 43.25          & 96.32\%         \\ 
SuperGlue\cite{superglue} & 52.05          & 22.75          & 90.94\%         & 57.04            & 80.44          & 96.61\%         \\ 
PCA-GM\cite{pcagm}    & 41.48          & 13.65          & 90.27\%         & 55.81            & 50.70          & 99.43\%         \\ 
FreeAlign & \textbf{0.017} & \textbf{0.266} & \textbf{0.56\%} & \textbf{0.015} & \textbf{0.318} & \textbf{0.46\%}   \\ \hline
\end{tabular}}
\vspace{-5mm}
\label{table:match}
\end{table*}

\subsection{Multi-Anchor Based Subgraph Searching}
\label{sec:multi}
Upon completion of the salient-object graph learning phase, the task is to locate the approximate maximum common subgraph between two salient-object graphs, signifying distinct and similar geometric structures. Let some potentially matched nodes pairs be anchors.
The core concept is that nodes are deemed a match if all edges linking the node and anchors in both graphs are similar, thereby forming a stable common subgraph with all anchors within its graph. Let $\mathcal{S}$ be the common subgraph and $\varepsilon$ be the confident score for this common subgraph, finding common subgraph between two salient-object graph $\mathcal{G}_i, \mathcal{G}_j$ whose edge feature tensors are $\mathbf{W}_i, \mathbf{W}_j$ by MASS can be formulated as
\begin{equation}
\small
    \mathcal{S}, \varepsilon = f_{\text{MASS}}\left(\mathcal{G}_i, \mathbf{W}_i, \mathcal{G}_j, \mathbf{W}_j\right).
\end{equation}

When there are $n$ and $m$ nodes in the $\mathcal{G}_i$ and $\mathcal{G}_j$ repectively. In order to realize $f_{\text{MASS}}\left(\cdot\right)$,  the procedure is divided into four primary steps:



\textbf{i). Initialization}. We begin by generating potential anchor lists. This involves $n\times m$ pairs of potential matching nodes to serve as the initial anchor for subsequent searches.

\textbf{ii). Anchor lists expanding}. For each initialized matched pair $(p,q)$ among the $n\times m$ potential node pairs, there is only one pair of anchors, which easily leads to instability given the various situations. To mitigate this, we expand the anchor lists. By examining another node pair $(u,v)$ and comparing the edge feature $\mathbf{W}_i(p,u)$ and $\mathbf{W}_j(q,v)$, we can ascertain the similarity. If the difference, denoted as $\varepsilon_{(p,u),(q,v)} = \vert \mathbf{W}_i(p,u)-\mathbf{W}_j(q,v)\vert$ is below a predefined threshold, then the pair $(u,v)$ will be added into the anchor list. is incorporated into the anchor list. This expansion proceeds until the count of anchor pairs meets the limit $\gamma$ ensuring that edge features related to all anchors are assessed. 

\textbf{iii). Subgraph search}. For every potential anchor list, nodes are incrementally added to the subgraph $\mathcal{S}$ until no other nodes meet the inclusion criteria.

\textbf{iv). Selection}. The discrepancy value, $\varepsilon$, is computed as $\varepsilon=\frac{1}{r^p}\sum\varepsilon_{e}$where $r$ represents the size of subgraph $\mathcal{S}$ and $p$ acts as a tunable hyperparameter. Out of all the subgraphs, the one with the minimal $\varepsilon$ is chosen to represent the common subgraph.

\subsection{Relative Transformation Calculation}

With MASS, we can get a common subgraph $\mathcal{S}$ and its score $\varepsilon$ between two salient-object graphs. To obtain the relative pose and time difference in \eqref{1b}, we need to calculate the relative transformation according to the common subgraph. It takes two step in temporal and spatial domain, respectively.

\textbf{Clock deviation estimation} To determine $\Delta t$, we leverage MASS between $\mathcal{G}_{j}^{t_{j\rightarrow i}}$ and a temporal salient-object graph buffer $[\mathcal{G}_{i}^{t_i}, \mathcal{G}_{i}^{t_i-\tau},..., \mathcal{G}_{i}^{t-l\tau}]$ in the $i$-th agent. After obtaining a list of matching results at different timestamps, the most reliable match is selected via the $\varepsilon$.
If a collaborative message fails to identify a common subgraph, whose number of nodes should exceed a predetermined minimum threshold, across the time buffer, \emph{FreeAlign} will discard this collaborative perception message to ensure safety. 

\textbf{Relative pose estimation.} To calculate $\Delta{\xi}_{j \rightarrow i}^t$, the common subgraph selected in temporal alignment is utilized to derive the relative pose $(\Delta x, \Delta y, \Delta \theta)$ as the common subgraph $\mathcal{S}_{ij}$ can be considered as two sets of matching points $\mathbf{S}_i^t, \mathbf{S}_j^t \in\mathbb{R}^{\psi \times 3}$ in two coordinate systems with robust RANSAC\cite{ransac} or LMesS\cite{lmeds} algorithm.

\subsection{Discussions}  

\textbf{Advantages.}  
Standard general alignment methods, such as SuperGlue\cite{superglue} and PCA-GM\cite{pcagm} cannot be applied to collaborative perception for two reasons: i) they leverage a keypoint detection network to extract key points as nodes with visual features, which are not provided in collaborative perception; ii) they leverage pixel coordinates to calculate edge attributes, which is highly affected by rotations. In comparison, \emph{FreeAlign} i) focus on geometric relationships of objects; and ii) leverages the relative distances, which are invariant from different perspectives.

\textbf{Prerequisites.} There are two assumptions for \emph{FreeAlign} to function well. i) Collaboration is initiated only when agents are in close proximity, ensuring a common field of view. 
ii) The scenario is dynamic. This makes it almost impossible for the geometric patterns at varying timestamps to be identical, allowing us to determine the timestamp. \textit{These assumptions are typically common and practical in a collaborative perception setting}.

\textbf{FreeAlign and GNSS Integration:} The \emph{FreeAlign} operates independently of GNSS inputs but can benefits from their integration. Firstly, it assesses the fundamental reliability of GNSS signal. Subsequently, GNSS enhances \emph{FreeAlign}'s efficiency by reducing its search range.

\section{Experimental Results}

\begin{table}
\captionof{table}{Collaborative detection under various latency deviations. \label{table:syncwithdetect}
\emph{FreeAlign} is robust to latency deviations.} \vspace{-1mm}
\centering
\small
\resizebox*{0.8\linewidth}{!}{
\begin{tabular}{l|cccc}
\hline
$\delta t_{j \rightarrow i}$(clock deviation)    & 0 & 100ms & 200ms & 300ms \\\hline
Single              & 0.78 &0.78 & 0.78 & 0.78 \\
Where2comm          & 0.66 &0.66 & 0.66 & 0.66 \\
Where2comm+SyncNet  & \textbf{0.87} &0.86 & 0.68 & 0.57 \\
Ours                & 0.86 &\textbf{0.86} & \textbf{0.86} & \textbf{0.86} \\\hline
\end{tabular}}
\vspace{-3mm}
\end{table}
\begin{table}[h]
\centering
\small
\captionof{table}{\label{table:sync}\emph{FreeAlign} deliver accurate temporal alignment in OPV2V and DAIR-V2X datasets.}
\resizebox*{0.8\linewidth}{!}{
\begin{tabular} {l|cc}
        \hline
                & Synchronization Accuracy & Average $\delta t_{j \rightarrow i}$ \\\hline
        OPV2V   & 79.46\%              & 22.8ms             \\
        DAIR-V2X & 66.36\%              & 45.6ms            \\\hline
\end{tabular}}
\vspace{-5mm}
\end{table}
\begin{table}[t]
\centering
\small
\caption{With \emph{FreeAlign}, all collaboration methods significantly improve their robustness to localization attack.} \vspace{-1mm}
\setlength\tabcolsep{3pt}
\renewcommand{\arraystretch}{1.2}
\resizebox*{0.99\linewidth}{!}{
\begin{tabular}[\linewidth]{l|ll|ll}
\hline
\centering
& \multicolumn{2}{c|}{OPV2V} & \multicolumn{2}{c}{DAIR-V2X}    \\ \cline{2-5}
          & AP@0.3       & AP@0.5 & AP@0.3     & AP@0.5      \\ \hline
Single & 0.79 & 0.78 & 0.70 & 0.66 \\ 
Late & 0.56 & 0.54  & 0.40 & 0.38 \\
V2VNet\cite{v2vnet} \textcolor{blue}{ECCV2020} & 0.44 & 0.40 & 0.42 &0.39 \\
DiscoNet\cite{disconet} \textcolor{blue}{NeurIPS2021} & 0.56 & 0.47 & 0.64 & 0.54 \\
V2X-ViT\cite{v2xvit} \textcolor{blue}{ECCV2022} & 0.60 & 0.59 & 0.55 & 0.52 \\
Where2comm\cite{where2comm} \textcolor{blue}{NeurIPS2022} & 0.32 & 0.30 & 0.46 & 0.44\\
CoAlign\cite{coalign} \textcolor{blue}{ICRA2023} & 0.32 & 0.30 & 0.64 & 0.61 \\\hline
V2VNet+FreeAlign & 0.88\textcolor{red}{$\uparrow 0.44$} & 0.75\textcolor{red}{$\uparrow 0.35$} & 0.69\textcolor{red}{$\uparrow 0.27$} & 0.60\textcolor{red}{$\uparrow 0.21$} \\
DiscoNet+FreeAlign & 0.87\textcolor{red}{$\uparrow 0.31$} & 0.79\textcolor{red}{$\uparrow 0.32$} & 0.76\textcolor{red}{$\uparrow 0.12$} & 0.70\textcolor{red}{$\uparrow 0.16$} \\
V2X-ViT+FreeAlign & 0.93\textcolor{red}{$\uparrow 0.33$} & 0.89\textcolor{red}{$\uparrow 0.30$} & 0.73\textcolor{red}{$\uparrow 0.18$} & 0.65\textcolor{red}{$\uparrow 0.13$} \\
Where2comm+FreeAlign & 0.86\textcolor{red}{$\uparrow 0.54$} & 0.71\textcolor{red}{$\uparrow 0.41$} & 0.75\textcolor{red}{$\uparrow 0.29$} & 0.67\textcolor{red}{$\uparrow 0.23$} \\
CoAlign+FreeAlign & \textbf{0.95\textcolor{red}{$\uparrow 0.63$}} & \textbf{0.94\textcolor{red}{$\uparrow 0.64$}} & \textbf{0.79\textcolor{red}{$\uparrow 0.15$}} & \textbf{0.73\textcolor{red}{$\uparrow 0.12$}}  \\ \hline
\end{tabular}}
\label{table:attack}
\vspace{-6mm}
\end{table}

\subsection{Datasets and Experimental Settings}
We conduct collaborative LiDAR-based 3D object detection on both 
a simulation dataset, OPV2V \cite{opv2v}, co-simulated by OpenCDA \cite{opencda} and Carla \cite{carla}, and a real-world dataset, DAIR-V2X \cite{dairv2x}. The detection results are evaluated by Average Precision (AP) at Intersection-over-Union (IoU) threshold of 0.30 and 0.50. We follow \cite{xu2022opv2v} and \cite{coalign} to set the detection range as $ x \in [-140m, 140m], y \in [-40m,40m]$ in OPV2V and $x \in [-100m, 100m], y \in [-40m,40m]$ in DAIR-V2X, respectively. We use PointPillars \cite{lang2019pointpillars} with the grid size $(0.4m,0.4m)$ as the encoder. For multi-scale feature fusion, the residual layer number is 3 and the channel numbers are $(64,128,256)$.

\begin{table*}[]
\centering
\small
\caption{Abalation studies on OPV2V and DAIR-V2X datasets.} \label{table:abalation}
\resizebox*{0.85\linewidth}{!}{%
\begin{tabular}{cc|ccc|ccc}
\hline
\multicolumn{2}{c|}{Dataset} & \multicolumn{3}{c|}{OPV2V} & \multicolumn{3}{c}{DAIR-V2X} \\
\hline
Anchor Based & GNN Feature & $\delta \theta$($^\circ$) & $\sqrt{(\delta x)^2 + (\delta y)^2}$(m) & Error Rate & $\delta \theta$($^\circ$) & $\sqrt{(\delta x)^2 + (\delta y)^2}$(m) & Error Rate \\
\hline
& & 4.657 & 0.979 & 3.71\% & 15.356 & 13.887 & 31.39\% \\
& \checkmark & 2.186 & 0.904 & 2.46\% & 5.721 & 4.962 & 9.59\% \\
\checkmark & & 0.029 & 0.283 & 0.78\% & 0.032 & 0.636 & 1.38\% \\
\checkmark & \checkmark & \textbf{0.017} & \textbf{0.266} & \textbf{0.56}\% & \textbf{0.015} & \textbf{0.318} & \textbf{0.46}\% \\
\hline
\end{tabular}}
\vspace{-2mm}
\end{table*}

\subsection{Quantitative Evaluation}
\textbf{Detection performance in presence of pose errors.} 
Fig. \ref{fig:sptial_result} compares detection performances of \emph{FreeAlign}'s  with the previous methods under varying pose noise levels on both OPV2V and DAIR-V2X datasets. The baselines include single-agent detection, V2VNet\cite{v2vnet}, DiscoNet\cite{disconet}, V2X-ViT\cite{v2xvit}, Where2comm\cite{where2comm}, CoAlign\cite{coalign} and late fusion. Gaussian noise $\mathcal{N}(0,\sigma)$ is applied to $(x,y,\theta)$, with $\sigma \in \left[0,8\right]$. We can see that \emph{FreeAlign}'s performance is robust, almost unaffected by pose noises, while the baseline methods show significant performance drops with pose noises. \emph{FreeAlign} improves detection by 20.5\% compared to single-agent perception under serious pose noises.

\begin{figure*}
  \centering
    \begin{subfigure}{.28\textwidth}
      \includegraphics[width=\linewidth]{ 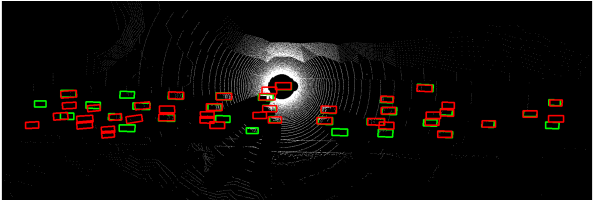}  
      \caption{V2X-ViT}
      \label{fig:sub-first}
    \end{subfigure}
    \begin{subfigure}{.28\textwidth}
      \includegraphics[width=\linewidth]{ 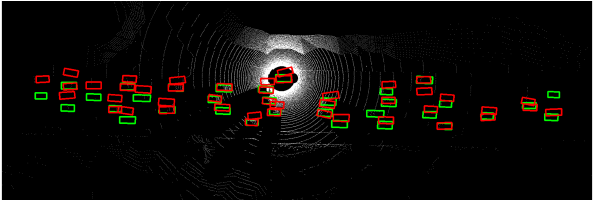}  
      \caption{CoAlign}
      \label{fig:sub-second}
    \end{subfigure}
    \begin{subfigure}{.28\textwidth}
      \includegraphics[width=\linewidth]{ 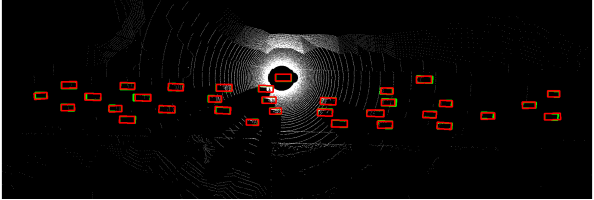}  
      \caption{FreeAlign}
      \label{fig:sub-second}
    \end{subfigure}
    \vspace{-2mm}
\caption{\emph{FreeAlign} qualitatively outperforms V2X-ViT and CoAlign on OPV2V dataset under pose noisy setting. \textcolor{green}{Green} and \textcolor{red}{red} boxes denote ground-truth and detection, respectively. }
\vspace{-3mm}
\label{fig:detectionresult}
\end{figure*}

\begin{figure*}[t]
    \centering
    \begin{subfigure}[b]{0.21\textwidth}
        \includegraphics[width=\textwidth]{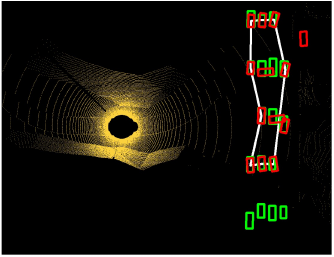}
        \caption{Ego view \\ \ }
        \label{fig:egoview}
    \end{subfigure}
    \begin{subfigure}[b]{0.21\textwidth}
        \includegraphics[width=\textwidth]{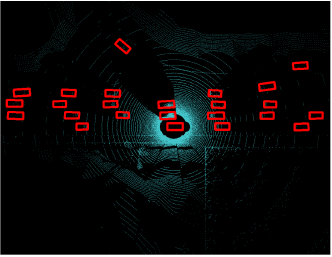}
        \caption{Temporal unmatched collaborator's view}
        \label{fig:edgeviewdelay}
    \end{subfigure}
    \begin{subfigure}[b]{0.21\textwidth}
        \includegraphics[width=\textwidth]{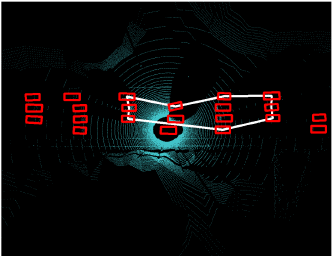}
        \caption{Temporal matched collaborator's view}
        \label{fig:edgeview}
    \end{subfigure}
    \begin{subfigure}[b]{0.21\textwidth}
        \includegraphics[width=\textwidth]{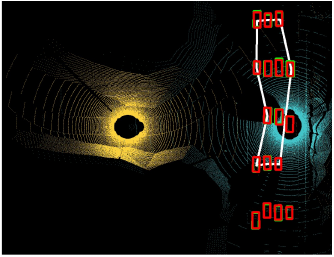}
        \caption{Fused ego view\\ \ }
        \label{fig:fuse}
    \end{subfigure}
    \caption{Visualization of collaboration between ego vehicle (a) and edge vehicle (b-c), and the spatial-temporal fusion result of them (d). \emph{FreeAlign} achieves spatial-temporal alignment through a common subgraph between views of different times and locations. \textcolor{green}{Green} boxes denote ground-truth, \textcolor{red}{red} boxes denote detection, \textcolor{yellow}{Yellow} and \textcolor{blue}{blue} denote the point clouds collected from ego and edge vehicles, respectively. White denotes the common subgraph.}
    \vspace{-5mm}
    \label{fig:qualitative}  
\end{figure*}

\textbf{Detection performance in the presence of latency deviation.}
To test~\emph{FreeAlign} in the presence of latency deviation (only know a noisy latency value), Table \ref{table:syncwithdetect} displays the AP 0.5 metric for the OPV2V dataset under 400ms latency with a noisy estimation, which means the compensation model only compensate for an inaccurate latency. The results show that i) without latency compensation, Where2comm's performance falls below that of a single agent; ii) SyncNet can compensate given an accurate latency value, but experiences a performance decline with increasing deviation; iii) our methods, Where2comm+SyncNet+FreeAlign can estimate a more accurate latency value, resulting in a notable gain over single-agent detection. The performances of single detection and Where2comm remain constant regardless of latency values as they do not compensate for latency. With the same setting, Table \ref{table:sync} further exhibits that \emph{FreeAlign} achieves an accurate temporal alignment in both datasets. Note that i) Synchronization Accurate means the \emph{FreeAlign} find the correct timestamp and see Table~\ref{notations} for the meaning of $\delta t_{j \rightarrow i}$. ii) there is no latency compensation module in Where2comm, so it will not be affected by clock deviation.

\textbf{Detection performance under localization attack.} 
We also explore detection performance under a malicious attack, a scenario where deliberately misleading pose data aims to generate false positive bounding boxes. Table~\ref{table:attack} presents the AP 0.3 and AP 0.5 of various methods under this attack. The results reveal that \emph{FreeAlign} significantly improves performance, enhancing CoAlign by 206.7\% on the OPV2V dataset, for instance. Notably, with \emph{FreeAlign}'s assistance, a majority of collaborative perception methods outperform single-agent perception even under malicious attack.

\textbf{Matching performance.}
To validate the performance on subgraph matching, we compare \emph{FreeAlign} with ICP\cite{icp}, SuperGlue\cite{superglue}, and PCA-GM\cite{pcagm}. Table \ref{table:match} shows the error between the estimated pose and the ground truth. Error rate means the proportion of cases that have transformation error greater than 3m (which makes an accurately detected box be regarded as false positive even in AP 0.3 metric).  We see that the other methods fail to match the corresponding subgraph and only the proposed \emph{FreeAlign} can achieve accurate matching in the collaborative perception scenarios.


\subsection{Ablation Studies and Discussions}
Table \ref{table:abalation} assesses the effectiveness of the proposed anchor-based matching and edge features learned by GNN. Absent the GNN features, edge matching is determined by the relative distance between two nodes. Without employing the anchor-based method, the search is anchored to a single pair. More results can be found in the supplementary material.
We find that 1) the anchor-based matching method is critical to our matching method. In DAIR-V2X dataset, without the multi-anchor setting, the error rate increases nearly 20 times; and 2) edge features learned by GNN help avoid failure cases, especially in a noisy real-world dataset DAIR-V2X. 
Results show the proposed anchor-based matching and GNN-based edge features make \emph{FreeAlign} more robust to noisy real-world applications.
\subsection{Qualitative Evaluations} 

\textbf{Visualization of detection results.}  
Fig. \ref{fig:detectionresult} shows a comparative visualization of detection results from V2X-ViT, CoAlign, and \emph{FreeAlign} in the OPV2V dataset under noisy setting. The noise stems from a Gaussian distribution with a standard deviation of 3.0m for position and 3.0° for heading. V2X-ViT, despite employing the MSWin module to mitigate pose error, struggles under large noise. Similarly, the pose graph optimization algorithm of CoAlign fails in the presence of large noise, leading to a significant drop in detection performance. In contrast, \emph{FreeAlign}'s exhibits superior performance under large noise. This can be attributed to its independence from prior pose information, which makes it less susceptible to the impacts of pose noise.

\textbf{Visualization of spatial-temporal alignment.}  
Fig.~\ref{fig:qualitative} offers a comprehensive visual depiction of \emph{FreeAlign}'s mechanism in a specific scenario, where the ego vehicle is located at a T-junction, limiting visibility. Fig. \ref{fig:egoview} displays the ego vehicle's perspective, illustrating suboptimal detection performance. Fig. \ref{fig:edgeviewdelay} displays the collaborator's perspective at an unsynchronized timestep, rendering it incompatible with the observation in Fig. \ref{fig:egoview}.
Fig. \ref{fig:edgeview} displays the collaborator's view at a synchronized timestep with ego. Note that there exist invariant geometric structures between Figs. \ref{fig:egoview} and \ref{fig:edgeview}. Fig. \ref{fig:fuse} demonstrates that, with the \emph{FreeAlign}'s assistance, the ego vehicle successfully detects through the T-junction.


\section{Conclusion}
This paper proposed \emph{FreeAlign}, a novel spatial-temporal alignment method for robust collaborative perception without any expensive high-end global localization and synchronized clock. The core idea is to leverage the invariant geometric structure composed of objects commonly perceived and observed by the individual agents to estimate the relative poses and time differences. Comprehensive experiments covering simulated and real-world datasets showed that \emph{FreeAlign} achieve efficient and accurate spatial-temporal alignment and help collaborative perception to be significantly more robust.

\clearpage



\bibliographystyle{IEEEtran}
\bibliography{IEEEfull}
\end{document}